%% file: root.tex
\documentclass[letterpaper, 10 pt, conference]{ieeeconf}  
\usepackage{graphicx}
\usepackage{mathtools}
\usepackage{amsmath}
\usepackage{xcolor}
\usepackage{float}
\usepackage{multirow}
\usepackage{array}
\usepackage{diagbox}
\usepackage{booktabs}
\usepackage{cite}

\usepackage{amsthm}

\theoremstyle{definition}

\theoremstyle{remark}

\usepackage{amssymb}

\IEEEoverridecommandlockouts                              

\newcommand*{\crosssymbol}{%
    \text{%
      \raise 1ex\hbox{%
        \rlap{\vrule height.2pt depth.2pt width .75ex}%
        \hbox to .75ex{\hss\vrule height .5ex depth 1ex\hss}%
      }%
    }%
}

\title{\LARGE \bf
Real-Time Model-Free Deep Reinforcement Learning for Force Control of a Series Elastic Actuator
}

\author{Ruturaj Sambhus$^{ \crosssymbol, *}$, Aydin Gokce$^{\crosssymbol}$, Stephen Welch, Connor W. Herron, and Alexander Leonessa
\thanks{\crosssymbol \,\,Joint first authors}
\thanks{*\,Corresponding author. E-mail: \tt\small ruturajsambhus@vt.edu}
\thanks{All authors are members of the Terrestrial Robotics Engineering and Controls (TREC) Lab,
        Virginia Tech, Blacksburg, VA 24060, USA.
        \{\tt\small ruturajsambhus, aydingokce, stephenwelchva, cwh, leonessa\} 
 \tt\small @vt.edu}%
}

\begin{document}

\maketitle
\thispagestyle{empty}
\pagestyle{empty}

\begin{abstract}
Many state-of-the art robotic applications utilize series elastic actuators (SEAs) with closed-loop force control to achieve complex tasks such as walking, lifting, and manipulation. Model-free PID control methods are more prone to instability due to nonlinearities in the SEA where cascaded model-based robust controllers can remove these effects to achieve stable force control. However, these model-based methods require detailed investigations to characterize the system accurately. Deep reinforcement learning (DRL) has proved to be an effective model-free method for continuous control tasks, where few works deal with hardware learning. This paper describes the training process of a DRL policy on hardware of an SEA pendulum system for tracking force control trajectories from 0.05 - 0.35 Hz at 50 N amplitude using the Proximal Policy Optimization (PPO) algorithm. Safety mechanisms are developed and utilized for training the policy for 12 hours (overnight) without an operator present within the full 21 hours training period. The tracking performance is evaluated showing improvements of $25$ N in mean absolute error when comparing the first 18 min. of training to the full 21 hours for a 50 N amplitude, 0.1 Hz sinusoid desired force trajectory. Finally, the DRL policy exhibits better tracking and stability margins when compared to a model-free PID controller for a 50 N chirp force trajectory.

\end{abstract}


\input{Sections/Section_1_Introduction.tex}
\newpage


\input{Sections/Section_2_System_Overview_and_Approach.tex}

\input{Sections/Section_3_Experimental_Setup.tex}
\input{Sections/Section_4_Results_and_Discussion.tex}

\input{Sections/Section_5_Conclusions.tex}
\bibliographystyle{ieeetr}

\bibliography{ref}

\end{document}

%% file: Sections/Section_1_Introduction.tex
\section{INTRODUCTION}






Series elastic actuators (SEA) are widely utilized for a variety of modern robotics applications such as humanoids \cite{hopkins2015embedded}, quadrupeds \cite{seadrl3}, exoskeletons \cite{herron2023design, searehab1}, and rehabilitation devices \cite{searehab2}. SEAs have numerous advantages such as low mechanical impedance, capacity to withstand impacts, improved stability bandwidth, and energy storage \cite{seapaper}. The performance of these robots relies on the stable force control capabilities of SEAs.

A simple control law for SEAs is a model-free PID controller \cite{conpid1}. While effective for ideal actuators, PID controllers alone present stability issues in the presence of nonlinearities such as stiction and backlash \cite{hopkins2015embedded, paine2013design}. The addition of model-based disturbance observer (DOB) to the PID controller uses a characterization of the ideal system to identify and eliminate nonlinearities for stable force control \cite{hopkins2015embedded, paine2013design}. In another approach, adaptive control can account for the model uncertainties and nonlinearities, but can be challenging to tune on hardware \cite{conadap1}. The mentioned model-based techniques require a detailed investigation into modeling the system accurately. This requires a strong understanding of the system or the use of meticulous system identification techniques to decouple the fundamental dynamics from the nonlinearities \cite{hopkins2015embedded, paine2013design}. Model-free learning-based approaches offer the ability to identify these key dynamics and achieve tracking objectives without even requiring an operator to be present. 

Recently, deep reinforcement learning (DRL) has been utilized to successfully solve a variety of high dimensional and non-linear control problems in simulation in a model-free manner \cite{drl2}. The works \cite{traj1,path1} used DRL to perform trajectory and path tracking of soft robotic manipulators, even though the agent was trained on a model in simulation. In the field of SEAs, DRL has been used for position control of low-cost SEAs in simulation \cite{seadrl2}, and locomotion of robots powered by SEAs \cite{seadrl3}. Few works in DRL are truly model-free to the extent that the policy is trained directly on the real robot. In \cite{seadrl3}, DRL is implemented directly on the real compliant quadruped to make an existing controller robust. As outlined in \cite{drllast,rllastplusone}, sim-to-real transfer works well on high-end commercial robots, but hardware under development may have noisy sensing, variable delays, non-stationarity, and mechanical non-idealities which are nontrivial to capture in a simulation. Hence, an on-hardware DRL framework would be valuable to obtain model-free control. 

To the authors knowledge, DRL methods have not been explored for force tracking of SEAs with force sensor-based feedback. As a step towards improving the force tracking of SEAs with non-linearities using model-free methods, the main contributions of this work are:


\begin{itemize}
  \item A real-time DRL framework for force tracking of a nonlinear SEA driving a pendulum which learns a policy entirely on hardware
  \item A detailed description of the environment setup and safety mechanisms for training on a hardware SEA system for 12 hours (overnight) without an operator
  \item Results showing improved tracking and stability margins of DRL policy compared to model-free PID

\end{itemize}
The paper is organized as follows: Section II describes the system overview and DRL approach, Section III discusses the experimental setup, Section IV details the results which demonstrate the efficacy of DRL in force control of a SEA compared to a model-free PID, and Section V is the conclusion.


%% file: Sections/Section_2_System_Overview_and_Approach.tex
\section{System Overview and Approach}

We propose a model-free DRL-based approach for force trajectory tracking of an SEA driving a linear pendulum without knowledge of dynamics. SEAs have non-linearities such as friction and backlash, and designing a traditional force feedback controller is nontrivial often requiring a system identification phase to develop a controller \cite{hopkins2015embedded, paine2013design}. The purpose of this work is to utilize a learning-based approach that learns a neural network control policy on-hardware to achieve stable force control of an SEA.

\subsection{Series Elastic Actuators}

\begin{figure}[t]
  \centering
  \includegraphics[width=1.0\linewidth]{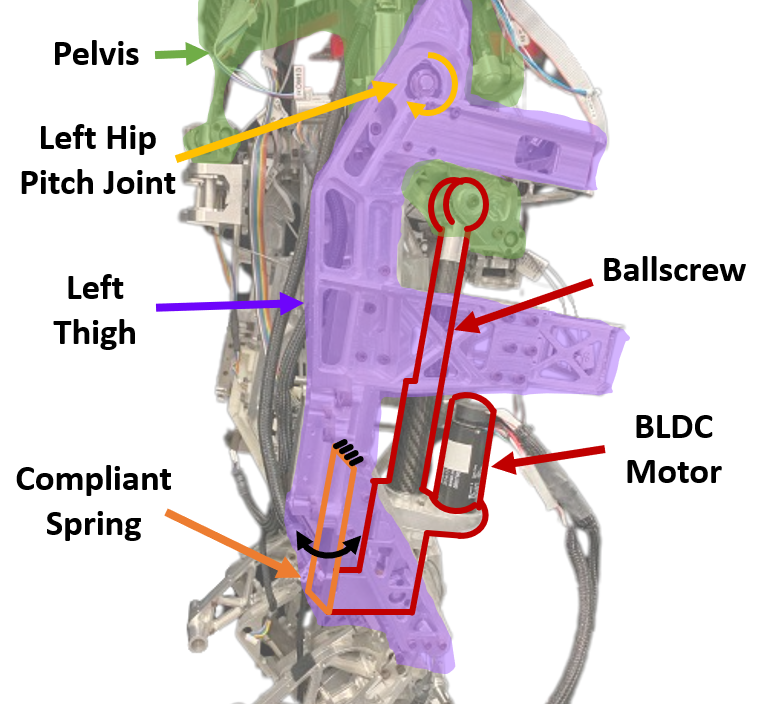}
  \caption{SEA utilized on ESCHER humanoid robot to drive the Left Hip Pitch Joint with mounting points on the Pelvis and Left Thigh.}
  \label{fig:sea_escher}
\end{figure}

SEAs are often modeled as second-order linear dynamical systems consisting of a motor with a gear train in series with a spring. As outlined in \cite{seapaper}, the spring-introduced compliance protects the actuators from jerk impacts that would otherwise damage the gears.
Often, the force control problem is turned into a position control (spring deflection) problem since the position is easier to control through the gear train. In this work, the SEA is designed with force sensor-based feedback, allowing for design flexibility when choosing the type of compliant mechanism. Utilizing elasticity in SEAs is crucial to ensure the stability of the force controller while limiting the overall bandwidth. Due to these properties, SEAs are widely used in human interactive systems such as haptic devices, prostheses, exoskeletons, and humanoid robots \cite{hopkins2015embedded,seaexo2, seadrl3, searehab1,searehab2,seahap1}. For example, as shown in Fig. \ref{fig:sea_escher}, an SEA is utilized for joint torque control to achieve stable, compliant whole-body behaviors even in the presence of disturbances such as uneven ground or a push. Traditionally, SEAs are heavily modeled using a system identification phase, which requires significant time investment and careful attention to detail to differentiate the core actuator dynamics from the test stand or any nonlinearities. Therefore, model-free approaches are desirable where an algorithm can explore the observation space and learn to control the actuator even without an operator present.

\subsection{Deep Reinforcement Learning}

An example of a model-free approach is Reinforcement learning (RL), which consists of an agent interacting with an environment modeled as a Markov Decision Process (MDP) \cite{sutton2018reinforcement}. In an MDP, an agent at time $t$ with state, $s_t\in S$, is capable of taking an action, $a_t\in A$, resulting in a reward, $r_t\in R$, where the next state, $s_{t+1} \in  S$, is governed by the state transition probability, $p(s_{t+1},r_t|s_t, a_t)$. The agent's strategy of choosing appropriate actions as a function of its state is known as a policy, denoted by $\pi_t(a_t|s_t)$. The goal of the reinforcement learning problem is to find a policy that maximizes the expected return $G_t \doteq \mathbb{E}[\sum_{t=0}^T \gamma^t R(s_t,a_t)]$ over time, where $\gamma \in [0,1]$ is the discount factor that weighs the importance of immediate rewards against delayed rewards. In Deep RL (DRL), policy $\pi _\mathbf{w}$ is parameterized by a neural network with $\mathbf{w}$ being the learnable weight matrix. The control problem is modeled as having a continuous state and action space. Hence, the action at time $t$ and state $s$ is given by the policy as $\pi_\mathbf{w}(a_t|s_t)$. The policy network outputs a mean and variance of the normal distribution from which the action is sampled.

Proximal Policy Optimization (PPO) \cite{ppo} is a state-of-the-art DRL algorithm known for its stable convergence over a wide range of hyperparameters \cite{persistence}. PPO consists of an actor policy network that attempts to predict the best action, and a critic, which estimates the actor's competency. PPO is an on-policy algorithm, meaning that the learning update is performed using the samples generated by the current policy. The goal of this work is to demonstrate the effectiveness of DRL learning directly on real hardware with a minimum number of experiments for hyperparameter tuning. Hence, PPO was chosen for its reliability. PPO is based on constraining the policy update so that the new policy is not very different, characterized in terms of the Kullback-Leibler (KL) divergence. This is done by clipping the surrogate objective function as shown in (\ref{eqn:ppoloss}). 
\begin{equation}\label{eqn:ppoloss}
    L^{CLIP}(\mathbf{w}) = \mathbb{\hat{E}}_t[min(r_t(\mathbf{w})\hat{A}_t, clip(r_t(\mathbf{w}), 1-\epsilon, 1+\epsilon)\hat{A}_t)]
\end{equation}
In (\ref{eqn:ppoloss}), $r_t(\mathbf{w})= \frac{\pi_\mathbf{w}(a_t|s_t)}{\pi_{\mathbf{w}_{old}}(a_t|s_t)}$ denotes the probability ratio, $\hat{A}_t$ is the advantage value estimated by the critic network, and $\epsilon$ is a hyperparameter typically equal to $0.2$. This loss function is computationally simple and yet is found to perform well for a wide range of DRL problems.

\begin{figure*}[t]
  \centering
  \includegraphics[width=\linewidth]{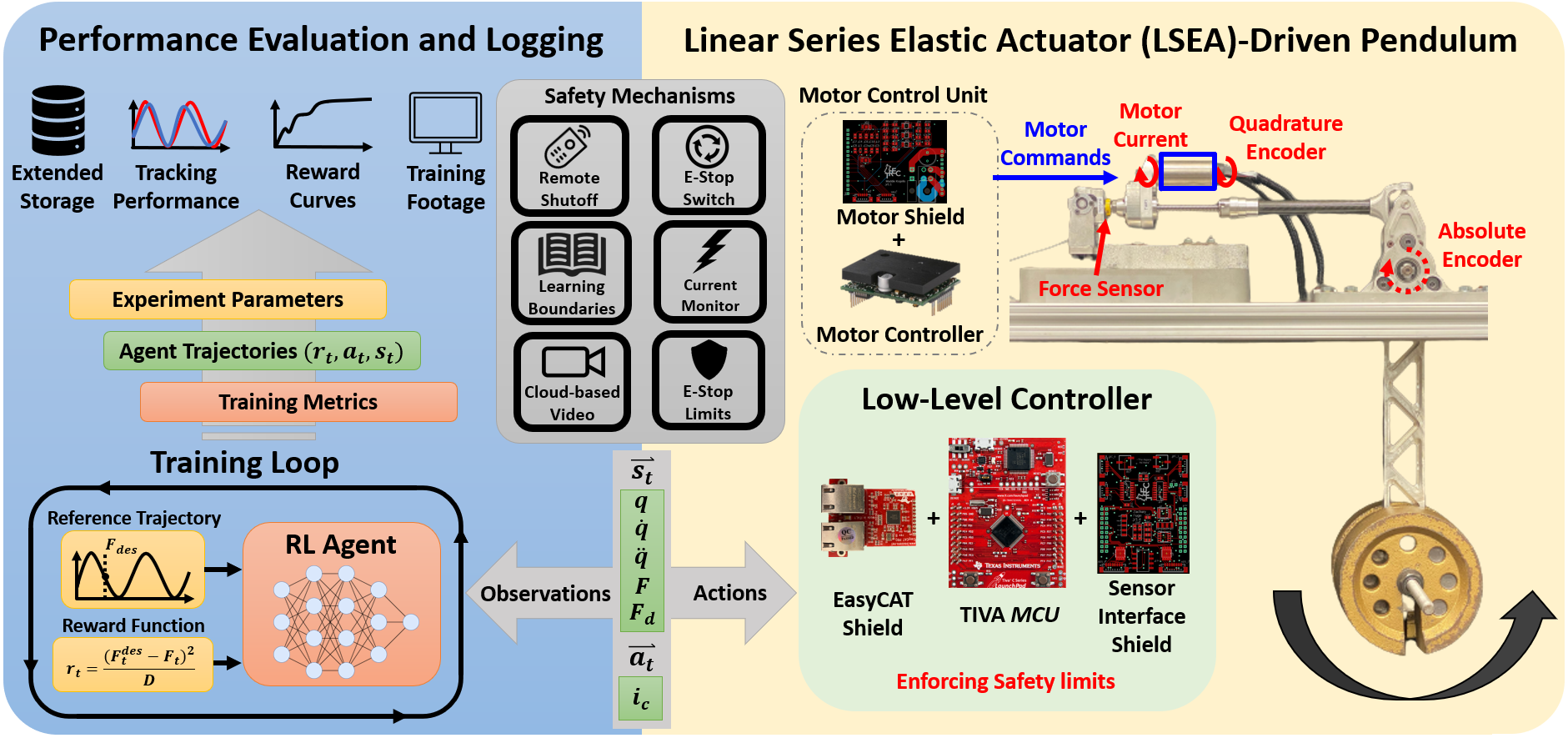}
  \caption{Linear SEA drives a pendulum with a 22.6 kg weight at the end of the lever arm. Low-Level Controller utilizes Motor Control Unit to send motor commands and networks the actions and observations with the RL Agent. Experiment parameters, agent trajectories, and training metrics are stored and evaluated to characterize policy performance. Safety mechanisms enabled the DRL policy to run for 12 hours (overnight) without an operator present.}
  \label{fig:experimental-setup}
\end{figure*}
\subsection{Learning on Hardware}

A common practice to train a DRL policy is to perform learning on a physics-engine simulator model of the system. In spite of the recent advances in high-fidelity physics engine simulation, the transfer of the learned policy to reality requires the use of many sim-to-real techniques detailed in \cite{sim2real}. SEAs may have unknown non-linearities with high-order dynamics which makes them significantly more difficult to accurately model. As discussed in \cite{hopkins2015embedded, paine2013design, seapaper}, SEAs are often modeled as ideal second-order linear systems for developing controllers and analyzing stability. For the SEA studied in this work, traditional system identification approaches revealed higher (+3) order dynamic properties which are nontrivial to represent in simulation. Full system identification is the focus of future work and is out of the scope of this paper. Overall, the traditional sim-to-real approach was not pursued, where the nonlinear effects such as stiction, backlash, and gear slipping could simply be encountered during on-hardware training.
When tuning a model-free PID controller, we observed that these nonlinear effects make the closed-loop system prone to instability. Therefore, reliable safety mechanisms must be developed for the hardware learning setup to autonomously train a DRL policy with minimal human supervision required. The details of the developed setup are mentioned in Section \ref{sec:ExerimentalSetup}. 

\subsection{Problem Formulation}

In this force trajectory tracking problem, the trajectories are restricted to sinusoidal waveforms between 0.05 to 0.35 Hz. This is because (as discussed in Section \ref{subsec:HardwareProtectedLearningMechanisms}) the input motor current is saturated to constrain the pendulum within safe position bounds and avoid any damage to the ball screw during learning. In addition, the initial random policy with abruptly switching high current values can induce vibration-based damage to the system. 

The assumption of the MDP does not hold on most real-world systems due to delays and hence those maybe only partially observable \cite{rllastplusone}. This problem is often countered by adding more information in the observation space of the agent. The observation vector includes the following: the absolute encoder-based angle, $q$ [rad], angular velocity, $\dot q$ [rad/s], and angular acceleration, $\Ddot{q}\, [rad/s^2]$, calculated by the forward Euler method, the current force measured, $F$ [N], and the desired force at the current time as per the trajectory, $F^{d}$ [N]. Hence, the observation space is a 5-dimensional vector. The $q$ is saturated to $\pm 0.25$ radians to maintain a safe margin to the physical boundaries of our system. The action space consists of a commanded current, $i_c$, saturated at $\pm0.75$ A. The reward function at time step $t$ is represented by the following,
\begin{equation} \label{eqn:reward}
    r_t=-\frac{(F^{d}_{t} - F_t)^2}{D},
\end{equation}
where the divisor, $D = 10^6$, maintains an episodic reward between [-30, 0]. In absence of this, the summation of reward per time step was observed to undergo catastrophic forgetting \cite{sutton2018reinforcement} resulting in unlearning and poor performance.


%% file: Sections/Section_3_Experimental_Setup.tex

\section{Experimental Setup}
\label{sec:ExerimentalSetup}


    


    
    
        
        


As displayed in Fig. \ref{fig:experimental-setup}, the linear SEA consists of a Maxon EC motor which belt-drives a THK ball screw through a gear transmission housing. The SEA is rigidly attached on the right end to a 0.3 m long pendulum with a 22.6 kg mass at the end. The pendulum's mass at the end was chosen to maximize the force amplitude within a smaller required range of motion of the joint angle to safely constrain the learning environment. The left end of the SEA is attached to a cantilevered elastic beam acting as a low-pass filter between the actuator and its environment. The development of the hardware and previous approaches of modeling and control of this SEA design can be found in \cite{knabe2014design, hopkins2015embedded}. 

As displayed in Fig. \ref{fig:experimental-setup}, there are several forms of sensor feedback for measuring the SEA state. An inline force sensor measures the applied actuator force which drives the pendulum arm. Two forms of encoder feedback are available including the motor quadrature (relative) encoder and joint (absolute) encoder. In addition, motor current feedback provides the applied torque from the motor onto the ball screw. For this paper, the motor encoder and current feedback are not utilized in the observation space, but may be integrated in future work. The actuator is controlled using a Low-Level Controller which is a hardware component made up of an EasyCAT networking shield, the Texas Instruments TIVA microcontroller, and an in-house designed Sensor Interface Shield. The Low-Level Controller communicates with a Networking PC using EtherCAT at 400 Hz. Further details about the hardware and networking approach can be found in \cite{herron2023design}.

\subsection{Experimental Setup Nonlinearities}

\begin{figure}[t]
      \centering
      \includegraphics[width=0.9\linewidth]{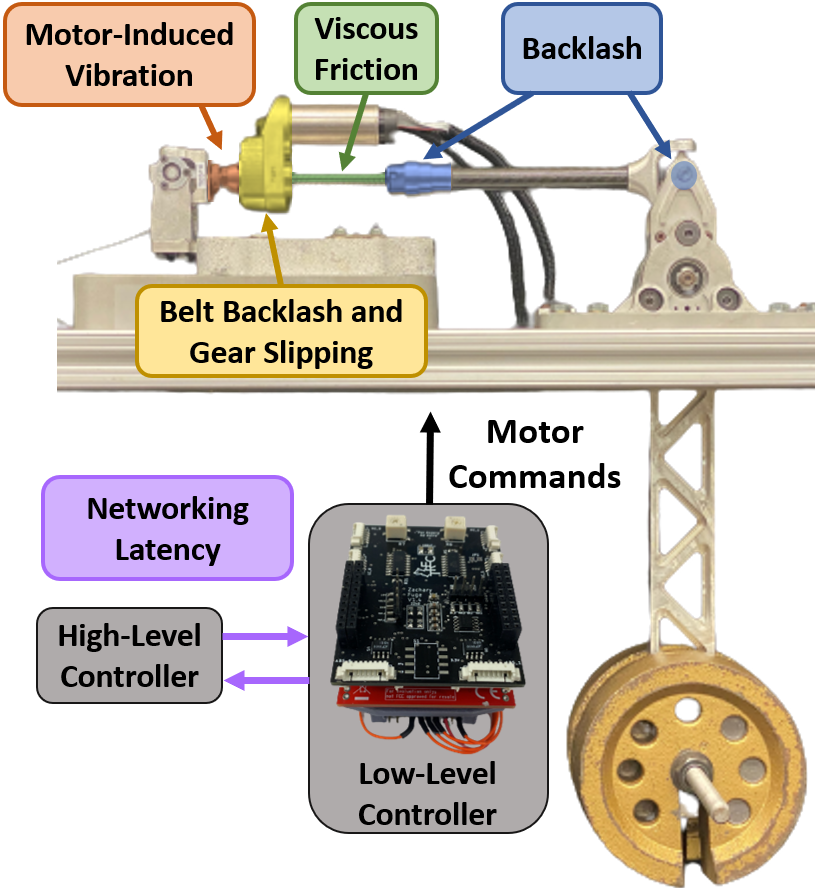}
      \caption{SEA and pendulum system contain several nonlinearities which can lead to instability without a robust controller or policy.}
      \label{fig:hardware-nonlinearities}
   \end{figure}

Series Elastic Actuators often contain several nonlinearities which directly limit the controller bandwidth and overall system stability. As displayed in Fig. \ref{fig:hardware-nonlinearities}, the SEA contains several mechanical nonlinearities such as viscous friction (stiction) along the ball screw, which induces a start-stop motion if unaccounted for. Within the ball screw/nut assembly, backlash may be present depending on the quality of the hardware. In addition, the actuator joint connection may also have backlash which creates large amplitude spikes in force measurement and discontinuous movement in the joint position measurement. Within the transmission housing, belt backlash and gear slippage can quickly lead to high, sharp command signals. Depending on the housing design and motor mounting strategy, motor-induced vibration can ruin a force signal at higher frequencies where the signal-to-noise ratio is nearly equivalent. Finally, networking latency between devices can create non-smooth command signals for a motor and delayed sensor feedback for computing the next input. Any of these nonlinearities in the mechanical, electrical, and communication subsystems can quickly lead to instability of the closed-loop. For SEA-driven systems, robust controllers are utilized to eliminate several of these unmodeled dynamic effects. The knowledge and time investment in developing an accurate model with system identification approaches for robust controllers motivates the need for an on-hardware learning-based strategy. However, an online learning strategy requires several levels of protection to ensure the safety of the hardware and allow the algorithm to fully explore the observation space without the need of an operator to reset testing.

\subsection{Hardware Protected Learning Mechanisms}
\label{subsec:HardwareProtectedLearningMechanisms}




The advantage of learning-based approaches is fully realized when an operator does not need to be constantly present to meticulously examine a system or tune a controller to achieve the desired behavior. A realizable DRL system should be persistent, meaning the agent is capable of collecting data and operating with minimal supervision \cite{persistence}. This necessitates the safety of the hardware system as well as the ability to autonomously collect data over long periods of time during testing. Therefore, it is imperative to guarantee hardware protection during the learning process through several levels of redundancy. 

    \begin{figure*}[t]
      \centering
      \includegraphics[width=\linewidth]{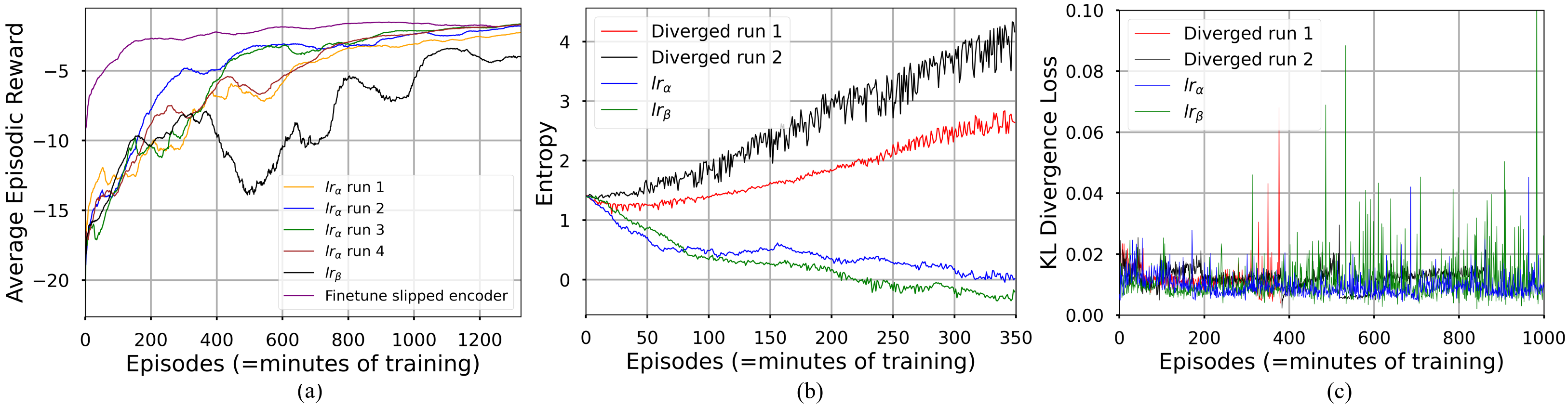}
      \caption{a) Learning curve for various learning rates (lr) decay schedules, where the reward is averaged over the number of training episodes completed b) Entropy plots used to choose lr schedule, c) KL Divergence Loss comparison of diverging experiments with the chosen lr decay ($l_\alpha$) and delayed lr decay ($lr_\beta$) schedule}
      \label{fig:reward_kl_div}
\end{figure*}

As displayed in Fig. \ref{fig:experimental-setup}, six safety mechanisms were utilized to constrain the algorithm even when left unsupervised for long periods of training and inference. The \textit{Emergency Stop Limits} for joint position ($\pm$ 0.35 rad) and actuator force constraints ($\pm$ 1700 N) are developed into the Low-Level Controller \cite{herron2023design} to shut off all motor commands if these limits are exceeded. While this ensures the safety of the device, it also would stop the learning process. Within the emergency stop joint position limits, a \textit{Learning Boundary} ($\pm$ 0.25 rad) allows the algorithm to freely command inputs within these joint boundaries. When the learning boundary has been exceeded and is still within the emergency stop limits, a constant current command is sent to move the pendulum away from the bounds towards the bottom of the swing curve. Generally, the value of opposing input can be regarded as a hyperparameter chosen in a way to avoid high-frequency oscillations at the bounds. With this bounding mechanism, we did not observe any violation of bounds over time as the policy improved. A DC power supply is the driving source for the motor and contains a \textit{Current Monitor} which limits the available range of command to $\pm$ 1.5 A. In conjunction, an \textit{Emergency Stop Switch} is onsite that directly severs the power source from the motor. Finally, a \textit{Cloud-based Video} feed allows an operator to monitor the learning performance while remote and a phone application allows for \textit{Remote Shutoff} of the system. A scenario where remote shutoff would be required is when the joint position encoder slips. In this case, the joint position no longer matches the intended model, emergency stop limits, and learning boundaries. Not only would this lead to unintended learning of the policy, but it could also damage the hardware. The best course of action is to shut off the power, recalibrate the joint encoder, and then reenable the learning process. These safety mechanisms enabled the DRL algorithm to run without an operator present for 12 hours (overnight) during the experimentation.



\subsection{Training Setup}

As displayed in Fig. \ref{fig:experimental-setup}, the training setup can be divided into two main components: a high-level PC and the Linear SEA controlled by the Low-Level Controller. The high-level PC contains a 12 GB Nvidia GeForce RTX 3080 GPU to enable fast tensor computations for running the DRL algorithms. The high-level PC computes the desired current, and the Low-Level Controller commands this current to the motor. Between these devices, an intermediary networking device exchanges the UDP socket commands of the high-level PC to the EtherCAT frames for the Low-Level Controller.

We used the Ray RLlib framework \cite{rllib} for learning where the DRL environment uses RLlib's ExternalEnv API. We chose this over the OpenAI Gym API, because the control is inverted: the environment queries the algorithm for actions, rather than the algorithm driving the environment. This makes the training process straightforward, enabling the environment to poll actions as necessary, all in a single thread operating at $100$ Hz. Weights and Biases (W\&B) \cite{wandb} is utilized to log the metrics of training. Moreover, we used cloud-based video streaming to remotely monitor the safety of the device in the initial phases of experimentation.

%% file: Sections/Section_4_Results_and_Discussion.tex
\begin{table}[b]
  
\begin{center}
\begin{tabular}{ |c|c|c|c| } 
\hline
Hyperparameters & Values \\
\hline
Policy Network & [256, 256]  \\ 
Value Network & [256, 256] \\
Activation & tanh  \\ 
Discount Factor $\gamma$ & 0.99  \\ 
Train batch size& 512  \\ 
Batch mode & Complete episodes  \\ 
\multirow{3}{*}{Learning Rate Schedule} &Linear decay from 5e-5 \\ 
& to 5e-6 in 1280 episodes, \\
& upto 1e-7 in 2048 episodes\\

\hline
\end{tabular}
\caption{Hyper-parameters}
\label{table:hyperparams}
\end{center}
\end{table}
\section{Results and Discussion}
The proposed approach is utilized for training a DRL policy on a SEA driving a pendulum. The target force trajectory has an amplitude of $50$ N with the frequency randomly sampled from the interval $0.05-0.35$ Hz kept constant throughout the episode lasting for one minute. The DRL policy is queried by the environment at $100$ Hz with minimal packet loss. A number of experiments are conducted for different learning rate schedules. The tracking performance of the DRL policy over various frequencies along with improvement with training time is discussed. Finally, the performance of the DRL policy is compared to a model-free PID.






 \begin{figure*}[t]
      \centering
      \includegraphics[width=\linewidth]{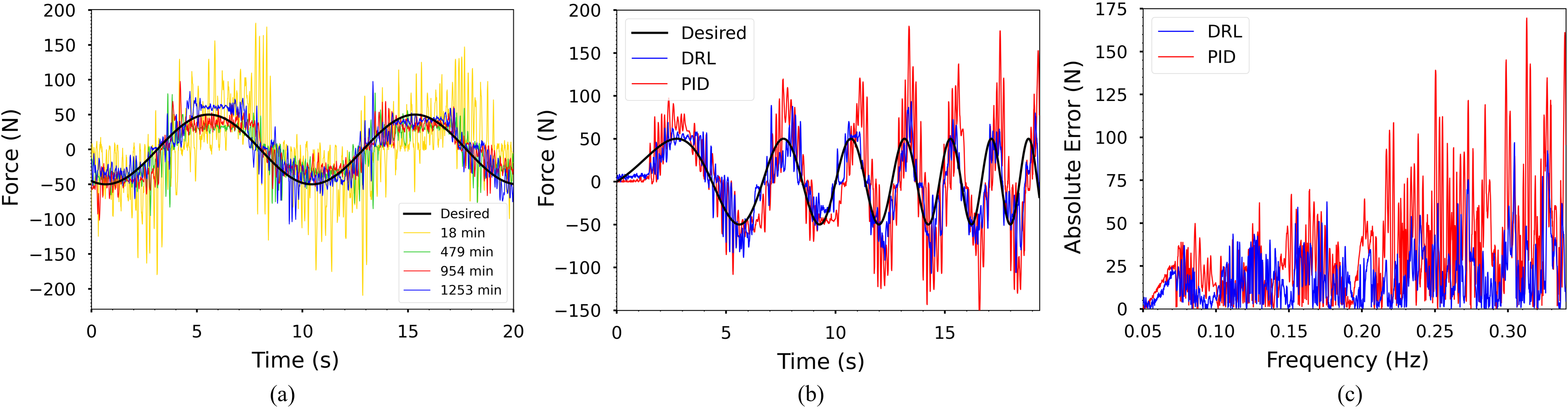}
      \caption{a) Comparison of 0.1 Hz desired trajectory displays improvements as training time increases, b) 50 N linear chirp signal (0.05-0.35 Hz) force tracking comparison between DRL and model-free PID, c) Absolute error comparison of DRL vs. PID for the 50 N linear chirp signal (0.05-0.35 Hz) force tracking}
      \label{fig:DRL_v_PID}
\end{figure*}

\subsection{Learning over time}

The RLlib agent is initialized with a random policy. As presented in Table \ref{table:hyperparams}, the policy and value networks are each chosen to be two-layer multilayer perception with $256$ neurons per layer. The default learning rate (lr) of $5e-5$ improved the reward, but the convergence was brittle since few experiments led to divergence. A higher lr was found to be ineffective towards any reward improvement. Observing the trends in entropy (regularization to promote agent exploration) and KL divergence loss (metric to quantify the difference between two policies), it was chosen to use a linear decay-based lr schedule. A delayed decay after $500$ episodes (denoted by $lr_\beta$) was found to degrade the reward due to the updated policy having a higher KL divergence loss, seen as green spikes in Fig. \ref{fig:reward_kl_div}c. Hence, the lr was higher leading to divergence from the optimal policy. From the experiments, an lr schedule (denoted by $lr_\alpha$) of $5e-5$ to $5e-6$ linearly in $1280$ episodes and further down to $1e-7$ linearly in $2048$ episodes was found to attain a stable convergence reliably, as in Fig. \ref{fig:reward_kl_div}a. None of the simulated benchmark hyperparameters for different environments by \cite{rllib}, \cite{SB3rl-zoo3} use a learning rate decay schedule. Hence, it would be valuable to note our findings about the importance of lr decay for hardware learning with delays, noise, and non-stationarity in sensing.



While experimenting with different hyperparameters and training configurations, the first $90$ minutes were crucial to predict the fate of a training run. In diverging experiments denoted as black and red in Fig. \ref{fig:reward_kl_div}b, the entropy consistently increased and the agent did not learn. If coarse trajectory tracking was not noticeable, the entropy consistently increased, or the KL divergence loss spiked beyond $0.02$ multiple times during the first $90$ minutes, the algorithm would be unlikely to converge with further training. In this work, which trains from scratch directly on the hardware, the 90-minute deadline for convergence was used as a pruning heuristic to save time.


For the policies which begin tracking within the first 90 minutes, training would be completed until the reward curve plateaus and starts oscillating with KL divergence loss spiking beyond the value of $0.02$, which would be approximately $1250$ episodes. Since each episode lasts a minute, this corresponds to approximately $1250$ minutes, or $21$ hours. As previously discussed in (\ref{eqn:reward}), the policy reward is based on force tracking performance.
As displayed in Fig. \ref{fig:reward_kl_div}a, the rate of reward improvement is found to be logarithmic with additional improvement requiring a longer time. This behavior can be seen in Fig. \ref{fig:DRL_v_PID}a, where the force tracking performance improves as the training time increases. The tracking after $18$ minutes of learning time is in phase with the target trajectory with a maximum overshoot of approximately $250 \%$. After $479$ minutes of training, the maximum overshoot was reduced to $50 \%$ of the desired trajectory amplitude with a small phase offset. The mean absolute error improved by about $60 \%$ relative to the $18$ minute mark. At the $1253$ minute mark, the reward additionally improved by about $7 \%$ after which any further improvement was found to be marginal. In addition, the instances of peak overshoot were reduced drastically when comparing $1253$ to $18$ minutes of training. Between 18 and 1253 minutes of training, the mean absolute error in force tracking for a 0.1 Hz sinusoidal force trajectory decreased by 25 N (63$\%$) and the average episodic reward over the number of trained episodes across 4 experiments improved by 13.4 (87 $\%$). 

A critical aspect of the DRL-based control approach is that it can generalize to changes in the system. Often during training, the pendulum angle encoder slipped due to vibrations in the system, where up to $0.1$ radians of slipping offset was observed over the course of a learning experiment. Since the observation space of the agent contains position, this nonlinear change in the system could significantly affect tracking performance and overall stability. A simple way to deal with slippage is to recalibrate the joint encoder several times during the course of the experiment. However, the DRL agent accommodated the offset during the learning and converged to equally good policies where encoder slippage was low. DRL has the potential to overcome the effect of non-stationarities in the sensing of state feedback quantities that do not contribute to the reward. A conventional controller using full-state feedback would be severely affected by such sensing offsets, which likely lead to instability. Further, the learned DRL policy for a particular offset could be re-trained on the recalibrated encoder with the learning curve (purple) shown in Fig. \ref{fig:reward_kl_div}a. The peak average reward was obtained in about $65 \%$ of the time ($900$ minutes) compared to learning from scratch. This robust approach could be beneficial for fine-tuning controllers to account for small changes such as degradation of parts.

\subsection{Comparison with PID controller}

An alternative model-free control approach is a PID controller, which is often the first baseline for stability and tracking performance due to its simplicity. Tuning a PID controller can often be difficult for systems with nonlinearities present and often lead to instability without heuristic modifications such as integral error and input saturation, gain scheduling, and filters. In this case, the PID tuning process was nontrivial often leading to instability even with the same input limits as the DRL policy. Proportional gain tuning often exhibited poor tracking or made the system violent with the force exceeding $1500$ N. The integral term would often lead to poor tracking or instability due to ballscrew stiction, where the controller would always move even while trying to achieve a constant desired force. Since the gains in the stable region did not provide satisfactory tracking, a first-order low-pass filter with a cutoff frequency of $1$ Hz on the error improved the stability. The filter cut-off frequencies above $1$ Hz were found to worsen the performance due to oscillations. 
After considerable PID tuning (10+ hours) 
over the sinusoidal target trajectory frequency range of $0.05-0.35$ Hz, the gains $K_p = 0.02$, $K_i = 0.002$, and $K_d = 0$ were chosen since $K_d$ was not found to affect the performance. 

It was important to note that despite the stability being sensitive to PID gains, the DRL policy was never unstable throughout the training. Even though there are no formal proofs of the stability of DRL policies, this policy had a $100 \%$ success rate with regard to stability without the use of any filtering. The learning boundaries on joint position aided in containing the algorithm to explore the observation space in a safe, stable manner. Even for training with hyperparameters that lead to reward degradation for $6+$ hours, the system still remained stable without triggering the emergency stop limits. This demonstrates the potential of learning to control custom-made mechanical hardware in development. An unstable controller can destroy such hardware in a fraction of a second. A DRL-based control policy can serve as a guiding policy to learn more about the dynamics of the system and collect data that could aid the design of a robust controller. More formal guarantees of the stability of DRL methods are necessary to compare and enhance robust control approaches.

   The results of the comparison with PID control are presented in Fig. \ref{fig:DRL_v_PID}b and \ref{fig:DRL_v_PID}c. It was observed that the tracking error increases as we increase the frequency of the target trajectory for both controllers. As displayed in Fig. \ref{fig:DRL_v_PID}c, the PID controller and DRL policy have similar absolute error values up to 0.15 Hz, while beyond this frequency the PID begins to significantly increase compared to DRL. This can be explained by the fact that the use of a low pass filter over the error for PID limits the bandwidth. The DRL policy does not have this limitation and hence performed better with increasing frequency up to $0.35$ Hz.
   
   As displayed in Fig. \ref{fig:DRL_v_PID}b, the tracking of both controllers at the lower frequencies exhibits a start-stop motion visible as square-shaped parts in the achieved trajectory. This can be explained by the presence of stiction nonlinearity \cite{hopkins2015embedded} in the SEA. At the beginning of the experiment when the frequency is at the lowest, the motion of the system with both controllers is slow due to the presence of stiction.

Although the DRL policy was trained in the frequency interval of 0.05 to 0.35 Hz, it was able to track the frequencies of $0.035$ and $0.5$ Hz, which lie outside the training frequency interval. This demonstrates that the tracking of the DRL policy is not just limited to the training frequencies. To achieve force tracking at higher frequencies, the saturated current limit must be increased. Learning from a random policy with high current leads to higher discontinuities in the input current causing violent vibrations. This can cause rapid wear of the mechanical parts with the possibility of damage. Hence, curriculum learning-based approaches \cite{curriculumlearning} need to be used where the current limits along with target force magnitude and frequency are gradually increased as the agent learns to solve the simpler problem. 

Designing a robust controller for the force control of SEA with nonlinearities such as stiction, and backlash becomes more challenging when driving a dynamic system with nonlinearities. Furthermore, modeling the actuator to account for unknown non-linearities, delays, and non-stationarities such as degradation over time would require investigating the system in great detail. With such systems, linear system identification methods may not capture the important dynamics. The DRL framework presented in this work would be very valuable to learn a black box controller even if it were for a limited action space. The data generated by a successful DRL controller would be valuable to learn about the system dynamics. To this end, the RL framework presented here is based on off-the-shelf tools available from open source without the need for explicitly engineering multi-processing, asynchronicity, delay-free communication, special modifications to the observation, action spaces, or the RL algorithm itself. The impact of the hyper-parameter of the learning rate decay schedule is highlighted for repeatable convergence. An example of the effect of non-stationarities in sensing and how DRL dealt with it is presented. Finally, the performance of the force tracking is not just competitive with PID but also safer in the stability margin. Additionally, robust control strategies utilize system identification to determine the core linear behavior, eliminating additional behavior from the expected plant as a disturbance \cite{hopkins2015embedded, paine2013design}. While this method is intuitive, the controller is static with respect to the desired plant dynamics and needs to be individually tuned for the joints on the robot. In future work, the achieved DRL controller will be benchmarked against a DOB controller and then enhanced by combining the two approaches where learning can help individually tune the disturbance dynamics for each joint on the robot.






%% file: Sections/Section_5_Conclusions.tex
\section{Conclusions}
A DRL policy based on PPO has been trained on hardware to achieve stable force control of an SEA driving a weighted pendulum with nonlinearities such as backlash, stiction, gear slippage, and latency. These nonlinearities would require nontrivial modeling decomposition to develop an accurate simulation for training a policy in a sim-to-real approach. Instead, this DRL policy was trained entirely on hardware with safety mechanisms that enabled learning to take place for 12 hours (overnight) without an operator present. The tracking performance improved by $25$ N in mean absolute error when comparing the policy after just 18 min to the full 21 hour training period for a 50 N, 0.1 Hz sinusoid force trajectory. When compared to a model-free PID controller, the DRL policy shows significant improvements in force tracking and stability. Future work will focus on comparing and integrating learning-based methods with existing robust control strategies for SEAs.

%% file: root.bbl
\begin{thebibliography}{10}

\bibitem{hopkins2015embedded}
M.~A. Hopkins, S.~A. Ressler, D.~F. Lahr, A.~Leonessa, and D.~W. Hong,
  ``Embedded joint-space control of a series elastic humanoid,'' in {\em 2015
  IEEE/RSJ International Conference on Intelligent Robots and Systems (IROS)},
  pp.~3358--3365, IEEE, 2015.

\bibitem{seadrl3}
A.~Raffin, D.~Seidel, J.~Kober, A.~Albu-Sch{\"a}ffer, J.~Silv{\'e}rio, and
  F.~Stulp, ``Learning to exploit elastic actuators for quadruped locomotion,''
  {\em arXiv preprint arXiv:2209.07171}, 2022.

\bibitem{herron2023design}
C.~W. Herron, Z.~J. Fuge, M.~Kogelis, N.~J. Tremaroli, B.~Kalita, and
  A.~Leonessa, ``Design and validation of a low-level controller for
  hierarchically controlled exoskeletons,'' {\em Sensors}, vol.~23, no.~2,
  p.~1014, 2023.

\bibitem{searehab1}
R.~Casas, T.~Chen, and P.~S. Lum, ``Comparison of two series elastic actuator
  designs incorporated into a shoulder exoskeleton,'' in {\em 2019 IEEE 16th
  International Conference on Rehabilitation Robotics (ICORR)}, pp.~317--322,
  2019.

\bibitem{searehab2}
H.~Yu, S.~Huang, G.~Chen, Y.~Pan, and Z.~Guo, ``Human--robot interaction
  control of rehabilitation robots with series elastic actuators,'' {\em IEEE
  Transactions on Robotics}, vol.~31, no.~5, pp.~1089--1100, 2015.

\bibitem{seapaper}
G.~Pratt and M.~Williamson, ``Series elastic actuators,'' in {\em Proceedings
  1995 IEEE/RSJ International Conference on Intelligent Robots and Systems.
  Human Robot Interaction and Cooperative Robots}, vol.~1, pp.~399--406 vol.1,
  1995.

\bibitem{conpid1}
E.~Garcia, J.~C. Arevalo, F.~Sanchez, J.~Sarria, and P.~Gonzalez-de Santos,
  ``Design and development of a biomimetic leg using hybrid actuators,'' in
  {\em 2011 IEEE/RSJ International Conference on Intelligent Robots and
  Systems}, pp.~1507--1512, IEEE, 2011.

\bibitem{paine2013design}
N.~Paine, S.~Oh, and L.~Sentis, ``Design and control considerations for
  high-performance series elastic actuators,'' {\em IEEE/ASME Transactions on
  Mechatronics}, vol.~19, no.~3, pp.~1080--1091, 2013.

\bibitem{conadap1}
D.~P. Losey, A.~Erwin, C.~G. McDonald, F.~Sergi, and M.~K. O’Malley, ``A
  time-domain approach to control of series elastic actuators: Adaptive torque
  and passivity-based impedance control,'' {\em IEEE/ASME Transactions on
  Mechatronics}, vol.~21, no.~4, pp.~2085--2096, 2016.

\bibitem{drl2}
Y.~Duan, X.~Chen, R.~Houthooft, J.~Schulman, and P.~Abbeel, ``Benchmarking deep
  reinforcement learning for continuous control,'' in {\em International
  conference on machine learning}, pp.~1329--1338, PMLR, 2016.

\bibitem{traj1}
A.~Robbins, M.~Ho, and M.~Teodorescu, ``Model-free dynamic control of robotic
  joints with integrated elastic ligaments,'' {\em Robotics and Autonomous
  Systems}, vol.~155, p.~104150, 2022.

\bibitem{path1}
S.~Satheeshbabu, N.~K. Uppalapati, T.~Fu, and G.~Krishnan, ``Continuous control
  of a soft continuum arm using deep reinforcement learning,'' in {\em 2020 3rd
  IEEE International Conference on Soft Robotics (RoboSoft)}, pp.~497--503,
  IEEE, 2020.

\bibitem{seadrl2}
F.~Sanfilippo, T.~M. Hua, and S.~Bos, ``A comparison between a two feedback
  control loop and a reinforcement learning algorithm for compliant low-cost
  series elastic actuators,'' in {\em Proceeding of the 53rd Hawaii
  International Conference on System Sciences (HICSS 2020)}, 2020.

\bibitem{drllast}
P.~B{\"o}hm, P.~Pounds, and A.~C. Chapman, ``Non-blocking asynchronous training
  for reinforcement learning in real-world environments,'' in {\em 2022
  IEEE/RSJ International Conference on Intelligent Robots and Systems (IROS)},
  pp.~10927--10934, IEEE, 2022.

\bibitem{rllastplusone}
G.~Dulac-Arnold, D.~Mankowitz, and T.~Hester, ``Challenges of real-world
  reinforcement learning,'' {\em arXiv preprint arXiv:1904.12901}, 2019.

\bibitem{seaexo2}
K.~Kong, J.~Bae, and M.~Tomizuka, ``A compact rotary series elastic actuator
  for human assistive systems,'' {\em IEEE/ASME transactions on mechatronics},
  vol.~17, no.~2, pp.~288--297, 2011.

\bibitem{seahap1}
D.~P. Losey and M.~K. O'Malley, ``Effects of discretization on the k-width of
  series elastic actuators,'' in {\em 2017 IEEE International Conference on
  Robotics and Automation (ICRA)}, pp.~421--426, IEEE, 2017.

\bibitem{sutton2018reinforcement}
R.~S. Sutton and A.~G. Barto, {\em Reinforcement learning: An introduction}.
\newblock MIT press, 2018.

\bibitem{ppo}
J.~Schulman, F.~Wolski, P.~Dhariwal, A.~Radford, and O.~Klimov, ``Proximal
  policy optimization algorithms,'' 2017.

\bibitem{persistence}
J.~Ibarz, J.~Tan, C.~Finn, M.~Kalakrishnan, P.~Pastor, and S.~Levine, ``How to
  train your robot with deep reinforcement learning; lessons we've learned,''
  {\em CoRR}, vol.~abs/2102.02915, 2021.

\bibitem{sim2real}
W.~Zhao, J.~P. Queralta, and T.~Westerlund, ``Sim-to-real transfer in deep
  reinforcement learning for robotics: a survey,'' {\em CoRR},
  vol.~abs/2009.13303, 2020.

\bibitem{knabe2014design}
C.~Knabe, B.~Lee, V.~Orekhov, and D.~Hong, ``Design of a compact, lightweight,
  electromechanical linear series elastic actuator,'' in {\em International
  Design Engineering Technical Conferences and Computers and Information in
  Engineering Conference}, vol.~46377, p.~V05BT08A014, American Society of
  Mechanical Engineers, 2014.

\bibitem{rllib}
E.~Liang, R.~Liaw, R.~Nishihara, P.~Moritz, R.~Fox, J.~Gonzalez, K.~Goldberg,
  and I.~Stoica, ``Ray rllib: {A} composable and scalable reinforcement
  learning library,'' {\em CoRR}, vol.~abs/1712.09381, 2017.

\bibitem{wandb}
L.~Biewald, ``Experiment tracking with weights and biases,'' 2020.
\newblock Software available from wandb.com.

\bibitem{SB3rl-zoo3}
A.~Raffin, ``Rl baselines3 zoo.'' https://github.com/DLR-RM/rl-baselines3-zoo,
  2020.

\bibitem{curriculumlearning}
S.~Narvekar, B.~Peng, M.~Leonetti, J.~Sinapov, M.~E. Taylor, and P.~Stone,
  ``Curriculum learning for reinforcement learning domains: {A} framework and
  survey,'' {\em CoRR}, vol.~abs/2003.04960, 2020.

\end{thebibliography}
